\documentclass[lettersize,journal]{IEEEtran}
\usepackage{amsmath,amsfonts}
\usepackage{algorithmic}
\usepackage{algorithm}
\usepackage{array}
\usepackage{textcomp}
\usepackage{stfloats}
\usepackage{url}
\usepackage{verbatim}
\usepackage{graphicx}
\usepackage{cite}

\usepackage{picinpar}
\usepackage[latin1]{inputenc}
\usepackage{colortbl}
\usepackage{soul}
\usepackage{multirow}
\usepackage{pifont}
\usepackage{color}
\usepackage{alltt}
\usepackage{enumerate}
\usepackage{siunitx}
\usepackage{epstopdf}
\usepackage{pbox}
\usepackage{amssymb}
\usepackage{mathtools}
\usepackage{bbm}
\usepackage{subfigure}
\usepackage{bbding}

\begin{document}

\title
{	
{LGSDF: Continual Global Learning of Signed Distance Fields Aided by Local Updating}
    
\thanks{This work is supported by the National Natural Science Foundation of China under Grant 62233002, 92370203. (Corresponding Author: Yufeng Yue, yueyufeng@bit.edu.cn)}
}

\author{
Yufeng Yue$^{1*}$, Yinan Deng$^1$, Jiahui Wang$^1$, and Yi Yang$^1$
\thanks{$^1$Yufeng Yue, Yinan Deng, Jiahui Wang,  Yi Yang  are with School of Automation, Beijing Institute of Technology, Beijing, 100081, China.}
\thanks{Manuscript received xxx.}
}



\maketitle

\begin{abstract}
\textcolor{black}{Implicit reconstruction of ESDF (Euclidean Signed Distance Field) involves training a neural network to regress the signed distance from any point to the nearest obstacle,} which has the advantages of lightweight storage and continuous querying. However, existing algorithms usually rely on conflicting raw observations as training data, resulting in poor map performance. In this paper, \textcolor{black}{we propose LGSDF, an ESDF continual Global learning algorithm aided by Local updating}. At the front end, axis-aligned grids are dynamically updated by pre-processed sensor observations, where incremental fusion alleviates estimation error caused by limited viewing directions. At the back end, a randomly initialized implicit ESDF neural network performs continual self-supervised learning guided by these grids to generate smooth and continuous maps. \textcolor{black}{The results on multiple scenes show that LGSDF can construct more accurate ESDF maps and meshes compared with SOTA (State Of The Art) explicit and implicit mapping algorithms. } The source code of LGSDF is publicly available at 
\urlstyle{tt}
\underline{\url{https://github.com/BIT-DYN/LGSDF}}.
\end{abstract}


\section{Introduction}
\IEEEPARstart{G}{enerating} an accurate and compact map of the surrounding environment incrementally in real-time is crucial for various downstream tasks, including autonomous robot exploration and virtual reality \cite{yue2020collaborative, eisentrager2023evaluating, leng2022efficient}. 
At present, there are many well-developed and widely applied explicit map structures, such as point clouds \cite{li2021rotation} and occupancy girds \cite{see-csom}. However, the map elements are discrete and the maintenance cost of these maps increases cubically with the resolution.
In recent years, implicit mapping \cite{deepsdf,nice-slam,yang2022recursive} has emerged as an effective solution to these problems, which utilizes the network with a simple architecture to reconstruct the scene. The network takes spatial 3D coordinates as input and outputs any environmental attributes required, such as density \cite{nerf}, occupancy \cite{occnet} and signed distance \cite{vox-surf}. In this paper, our work focuses on recovering the implicit ESDF (Euclidean Signed Distance Field) of the perceptual scene. ESDF has the advantage of directly providing signed distance and direction from the query point to the nearest obstacle, which can be seamlessly applied to path planning and trajectory optimization.

\begin{figure}[!t]\centering
    \subfigure[Real RGB Image]{\label{first_11}
    \includegraphics[width=4.25cm]{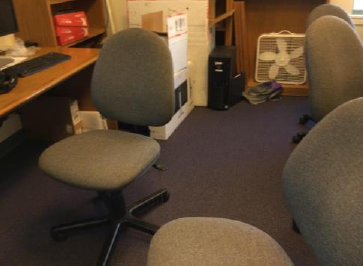}}
    \subfigure[SOTA ESDF Explicit Map]{\label{first_12}
    \includegraphics[width=4.25cm]{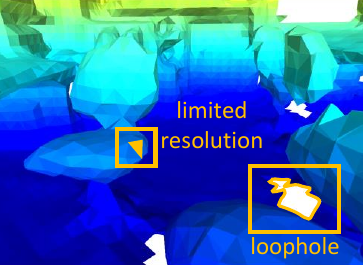}}
    \vfill
    \subfigure[SOTA ESDF Implicit Map]{\label{first_21}
    \includegraphics[width=4.25cm]{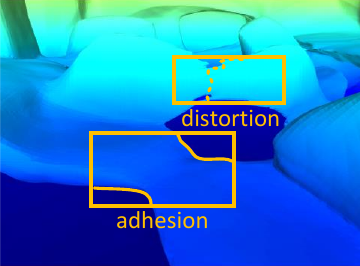}}
    \subfigure[LGSDF Map]{\label{first_22}
    \includegraphics[width=4.25cm]{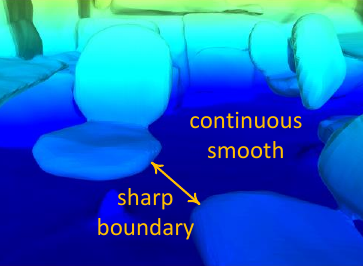}}
    \caption{The demonstration of zero-level set meshes of ESDF constructed by different strategies. Explicit mapping dynamically updates signed distances within the discrete grids and thus has limited resolution and is not predictive for unobserved regions. Implicit mapping regresses the SDF using a simple MLP, but often causes scene distortion due to conflicting training data. In contrast to these, LGSDF reconstructs a continuous, smooth map with sharp object boundaries.}
\label{first}
\end{figure}

The current methods for constructing ESDF maps can be broadly categorized into two main strategies. Explicit mapping \cite{voxblox} to construct ESDF generally involves fusing observations from depth sensors into a surface reconstruction of obstacles, and inferring the full SDF using a wavefront propagation algorithm. This approach offers a gridded perceptual space and direct 3D storage of signed distance values. However, grid-based local updates can only obtain limited-resolution details and cannot hallucinate unseen parts of the scene without a geometric context module (See \ref{first_12}). 
Unlike the intuitive representation of explicit mapping, \textcolor{black}{implicit mapping \cite{isdf} employs coordinates and approximate distances of the sampled points to train a MLP (Multilayer Perceptron) online as the requested map.} This approach offers predictive and fault-tolerant characteristics, allowing the SDF of the entire scene to be continuously regressed and smoothly denoised. However, distance approximation may differ at the same location due to different perceptual perspectives, leading to conflicts in training data and poor learning outcomes (See \ref{first_21}). Furthermore, efficiently addressing catastrophic forgetting in continual learning remains an important challenge.  Replay \cite{replay} is a popular approach but requires maintaining historical keyframes, increasing memory overhead, and exacerbating the impact of conflicting training data.

To address the above challenges, this paper proposes a novel online continual learning algorithm for building ESDF maps, called LGSDF. 
At the front end, LGSDF incorporates Active Sampling and Local Updating modules. Active Sampling is performed at both pixel and point levels to obtain 3D sampled points, \textcolor{black}{with the novel pixel sampling strategy assigning more attention to complex foreground instances.}
Local Updating approximates the signed distances to the nearest obstacle for these sampled points and fuses them into discrete axis-aligned grids.
At the back end, LGSDF includes the Global Learning module, where a randomly initialized neural network is trained in a self-supervised manner, guided by center coordinates and currently stored distances of updated grids. 
The grids selected for training consist of two parts, the currently updated grids and the sampled historically updated grids, ensuring the plasticity and stability of the network, respectively.
Through the successive operation between the front end and the back end, the network is continually optimized as an implicit ESDF map of the observed scene, achieving both sufficient local details and global continuity. Results on multiple public scenes verify the high accuracy and neatness of the proposed algorithm on both ESDF maps and meshes.
The contributions are listed below:

\begin{itemize}
    \item \textcolor{black}{We propose LGSDF, a novel ESDF implicit mapping algorithm, which for the first time uses local updating of the grids to aid in continual global learning of the neural implicit map.}
    \item \textcolor{black}{In the active sampling module, we devise a novel pixel sampling strategy to ensure that more perceptual attention is allocated to structurally complex foreground instances, thereby improving overall detail quality.}
    \item \textcolor{black}{In the global learning module, we design a novel grid (i.e., training data) selection strategy to ensure that the neural implicit map is both historically memorized and dynamically sensitive.}
 
\end{itemize}

The remainder of this paper proceeds as follows. Section \ref{RW} reviews the related works. Section \ref{LGSDF} introduces the LGSDF algorithm. Section \ref{ER} shows the experimental results. Section \ref{Conclusion} concludes the paper.

\section{Related Works} \label{RW}

In this section, existing related algorithms are introduced, including Explicit Reconstruction and Implicit Representation.

\subsection{Explicit Reconstruction}
Explicit mapping tends to store map information directly in a 3D discrete space.
Over the past decades, many explicit mapping frameworks have been employed for environment reconstruction \cite{hd-ccsom}. OctoMap \cite{OctoMap} is a highly popular and efficient occupancy mapping method, which utilizes a hierarchical tree structure to store occupancy. Meyer-D et al. \cite{Occupancy} present a probabilistic grid-based approach for modeling changing environments, storing both occupancy and its changes. S-MKI \cite{s-mki} builds a locally continuous occupancy map using a multi-entropy kernel function.
\textcolor{black}{In addition to the occupancy maps, TSDF (Truncated Signed Distance Function) \cite{TSDF_FUSION} is a common representation for recovering the surfaces.} It has been integrated into many online mapping frameworks, such as KinectFusion \cite{kinectfusion} and Bundlefusion \cite{bundlefusion}. However, these methods favor the construction of obstacles rather than traversable spaces, limiting the application scenarios.

As a complete representation of TSDF, ESDF can provide the full SDF, which is crucial for robot obstacle avoidance and motion planning. Due to the huge computational complexity, only a few algorithms can construct ESDF maps incrementally in real-time. For example, Voxblox \cite{voxblox} is a method to progressively build ESDF from TSDF based on the idea of wavefront. \textcolor{black}{Other related works, such as FIESTA \cite{FIESTA} and VDB-EDT \cite{vdb} perform Euclidean distance transform and the ESDF maps are derived from occupancy maps.} 
However, these methods only compute distances for discrete grids, so the accuracy of the maps is limited by the grid resolution, and they cannot predict the information about invisible regions due to the lack of a geometric context module. 
\textcolor{black}{Therefore, we proceed in the framework of implicit representation, using a network to implicitly define a nonlinear correspondence between arbitrary locations and signed distance values.}

\begin{figure*}[!t]\centering
	\includegraphics[width=18.2cm]{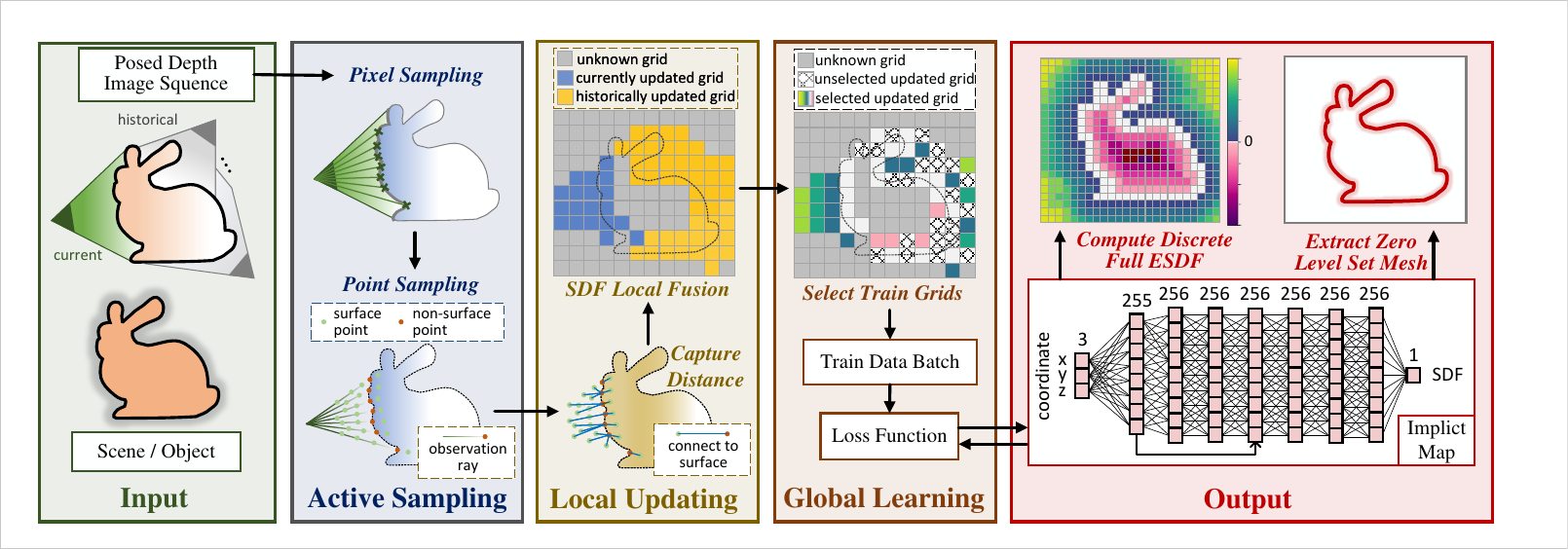}
	\caption{The framework of LGSDF consists of three main modules: Active Sampling, Local Updating and Global Learning. With the continuous input of the posed depth images flow, the implicit ESDF neural network gradually recovers the full SDF of the observed scene through self-supervised continual learning.} 
	\label{framework} 
\end{figure*}

\subsection{Implicit Representation}

In recent years, as NeRF \cite{nerf} demonstrated the feasibility of neural radiance fields, learning-based implicit mapping algorithms have sprung up intending to learn continuous implicit functions for 3D representations. This idea is originally applied to the monolithic reconstruction of objects, such as OccNet \cite{occnet} and Deepsdf \cite{deepsdf}. While offline unified object reconstruction is useful for computer vision tasks such as image rendering, online incremental environment reconstruction can better meet the practical needs of robots to explore unknown environments autonomously. 

The existing incremental implicit mapping can be further categorized into two types. \textcolor{black}{Although both adopt a learning-based approach, they differ in their selection of learning patterns. 
Algorithms following the first pattern store feature embeddings in voxel centers or corners and utilize a shared decoding network as a separate output module, such as RoutedFusion \cite{routedfusion}, DI-Fusion \cite{di-fusion}, NICE-SLAM \cite{nice-slam}, Go-surf \cite{go-surf} and SHINE-Mapping \cite{shine}.
Although performing continuity only within the voxel scale improves local accuracy but sacrifices global predictability. In addition, discrete queries and storage incur significant computational and storage costs.}

\textcolor{black}{Algorithms following the second pattern use a unique MLP to regress density, occupancy or signed distance at arbitrary locations, such as iMAP \cite{imap}, CNM \cite{cnm} and iSDF \cite{isdf}. The maps obtained in this way are globally continuous.}
ISDF is the first work to show that simple network models can be trained in a real-time continual learning setting as implicit representations of ESDF maps. However, it extracts training data from raw depth observations, which may conflict with each other, especially when perceived orientations vary widely. These conflicts will lead to overfitting and distortion of the final map.        

\textcolor{black}{Similar to iSDF \cite{isdf}, we continue to employ a straightforward MLP as the representation for the entire scene. However, our distinctive approach lies in the utilization of discrete grids, where raw sensor data is fused, as our training data. This method allows us to build ESDF maps characterized by local sharpness and accuracy, while simultaneously ensuring global smoothness and continuity.}

\section{SDF Global Learning Aided by Local Updating} \label{LGSDF}

This section describes and formulates the LGSDF algorithm, divided into four subsections: Algorithm Overview, Active Sampling, Local Updating and Global Learning.

\subsection{Algorithm Overview} \label{AO}

\subsubsection{Neural Implicit ESDF Map}

The Euclidean Signed Distance Field is a scalar field that associates the signed distance from each point in space to the nearest obstacle. Points within the obstacle are assigned minus signs to indicate unreachable regions, while those in free space are assigned positive signs to denote traversable regions. Such a 3D scalar field can be approximated by a neural network: 
\begin{equation}
\begin{aligned}
    \label{network}
    {{f}_{\theta}}:\underset{x,y,z}{\mathop{{{\mathbb{R}}^{3}}}}\,\to \underset{d}{\mathop{\mathbb{R}}}\,
\end{aligned}
\end{equation}
where $\theta$ is the network parameter, $\{x,y,z\}$ is the global 3D coordinates of the query location $\textbf{q}$, and $d$ is the predicted signed distance. The textured surfaces of obstacles can be extracted at those locations where the predicted distance is $0$, also known as zero-level set mesh:
\begin{equation}
\begin{aligned}
    \label{surface}
    \mathcal{S}=\{\textbf{q}\in {{\mathbb{R}}^{3}}|{{f}_{\theta }}(\textbf{q})=0\}
\end{aligned}
\end{equation}

The network structure is designed as a simple MLP with 6 hidden layers of feature size 256 (See Fig. \ref{framework}), and each intermediate layer is activated using Softplus. Before input to the network, the coordinate $\{x,y,z\}$ are converted into higher-dimensional vectors by applying `off-axis' positional embedding \cite{embedding}, which is crucial for reconstructing high-frequency signals. The embedded vectors are also concatenated to the third layer of the network. 

\subsubsection{Algorithm Framework}
\textcolor{black}{LGSDF takes a sequence of posed depth images as input and continuously trains a randomly initialized neural network as the final implicit ESDF map. The depth images are captured using a depth camera, and their corresponding camera poses are typically obtained by off-the-shelf RGB-D localization techniques.} 

The framework of the LGSDF algorithm is depicted in Fig. \ref{framework}. 
LGSDF consists of three main modules, the first two of which belong to the front end and the last one to the back end. \textcolor{black}{Firstly, the Active Sampling module selects 3D points from the viewing frustum to reduce computational complexity. These points are then fed into the Local Updating module to capture the signed distance from each point to the nearest obstacle and dynamically update the discrete grids. Finally, the Global Learning module utilizes these updated grids to supervise the learning process of the implicit ESDF neural network. For each new incoming frame, these three modules are sequentially executed, ensuring that the implicit map always reflects the most up-to-date results.} As the posed depth images are continuously input, the network gradually restores the full SDF of the observed scene. We visualize the output in two forms, i.e., the discrete ESDF map of arbitrary resolution and the mesh extracted using Marching Cube \cite{marching_cubes}.

\subsection{Active Sampling}

It is computationally infeasible and redundant to update grids with a dense point cloud within the viewing frustum of input images.  Therefore, the active sampling module is the first to be designed to select the more critical data.
As in previous works \cite{imap, isdf}, a complete active sampling procedure is usually executed sequentially at three levels: frame level, pixel level and point level. 
One of the main functions of frame level sampling is to select some keyframes, serving as a replay memory bank to avoid network forgetting.
However, this mechanism is not necessary for LGSDF, \textcolor{black}{because the regular distribution of training data (i.e., grids) makes it possible to preserve historical memory without replaying. Instead, LGSDF consistently processes the most recent captured frame to ensure its online real-time processing capability.  } 

\subsubsection{Pixel Sampling}
\begin{figure}[!t]\centering
    \subfigure[Depth Image]{\label{sampling_11}
    \includegraphics[width=4.25cm]{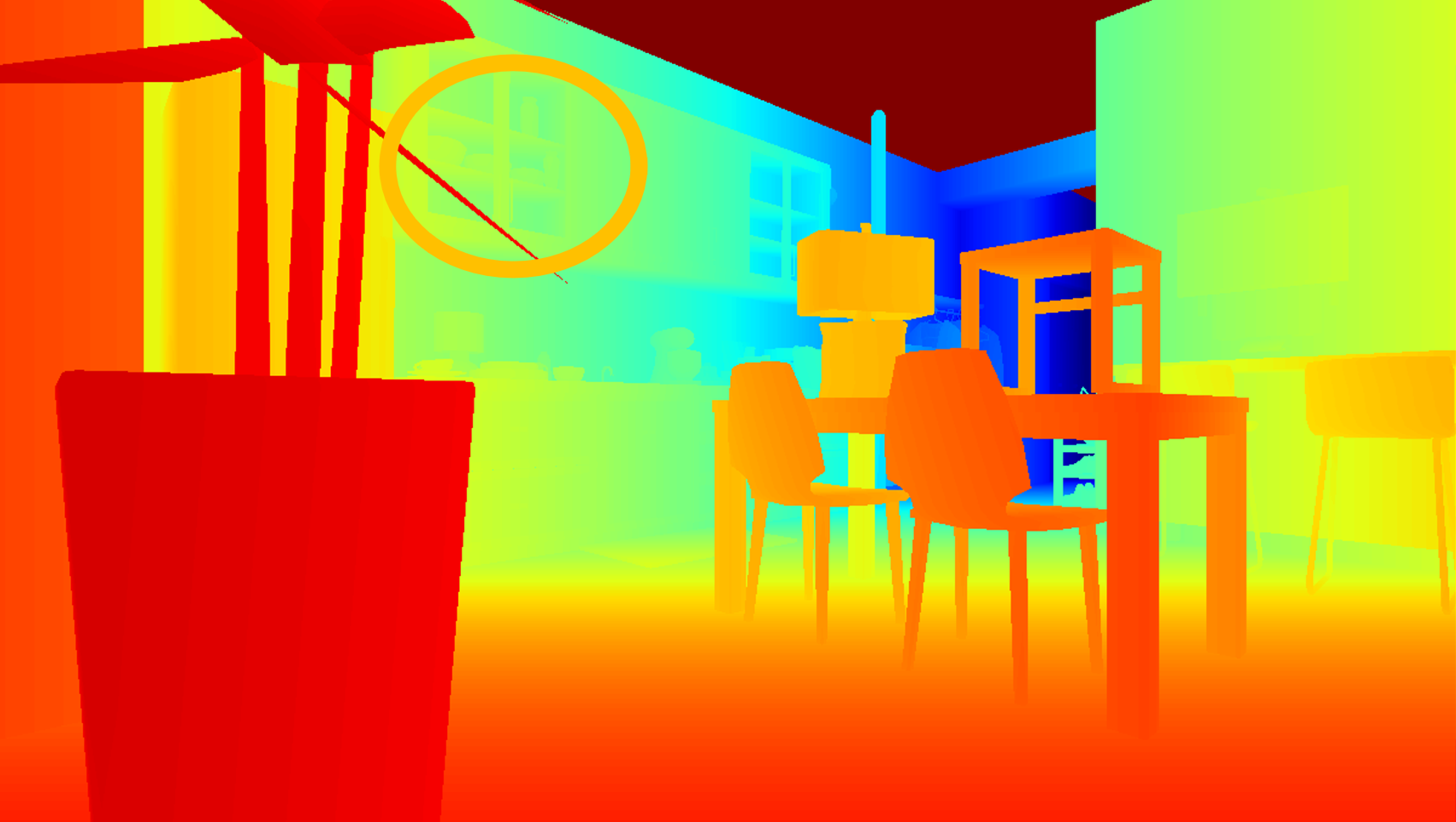}}
    \subfigure[Normal Rendering Image]{\label{sampling_12}
    \includegraphics[width=4.25cm]{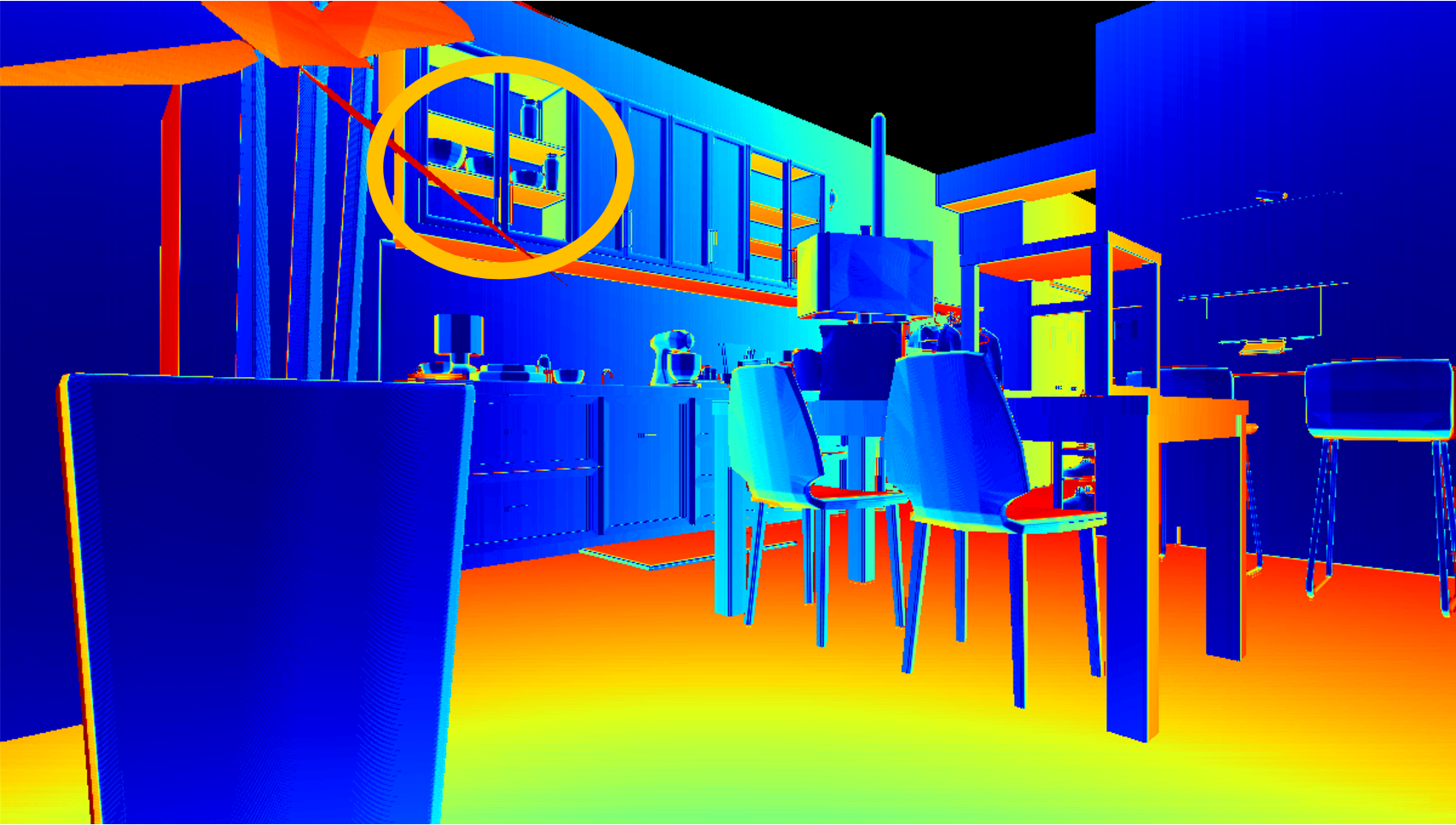}}
    \vfill
    \subfigure[Normalized Irregularity]{\label{sampling_21}
    \includegraphics[width=4.25cm]{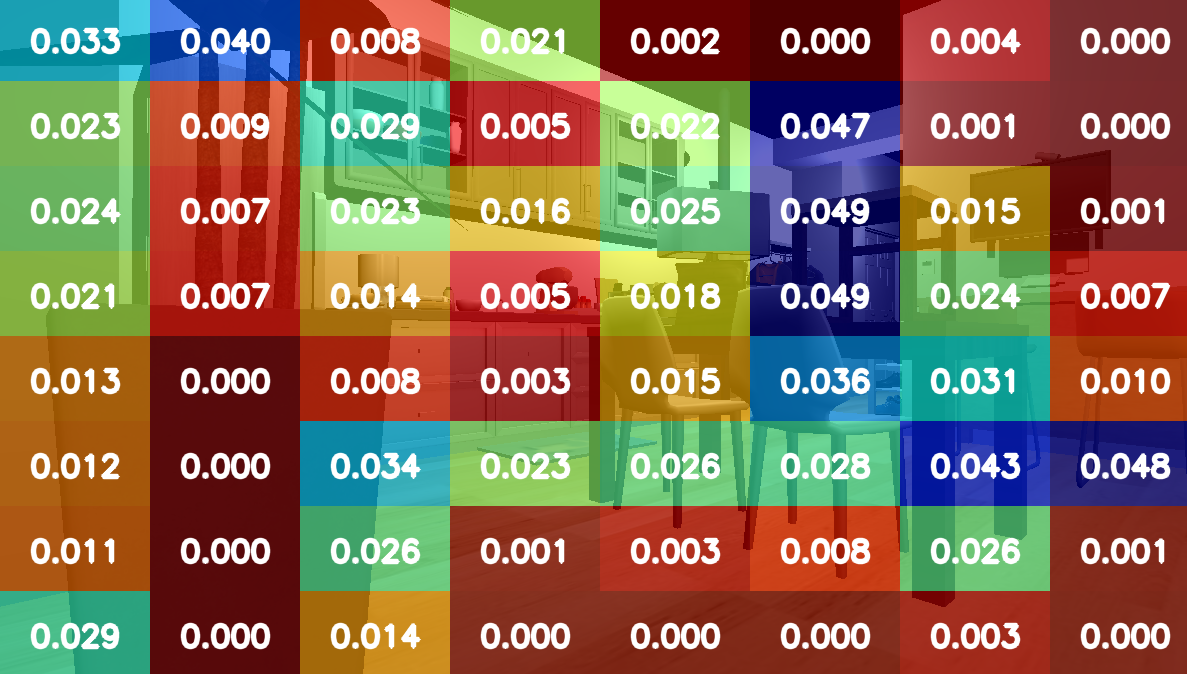}}
    \subfigure[Sampling Reuslt]{\label{sampling_22}
    \includegraphics[width=4.25cm]{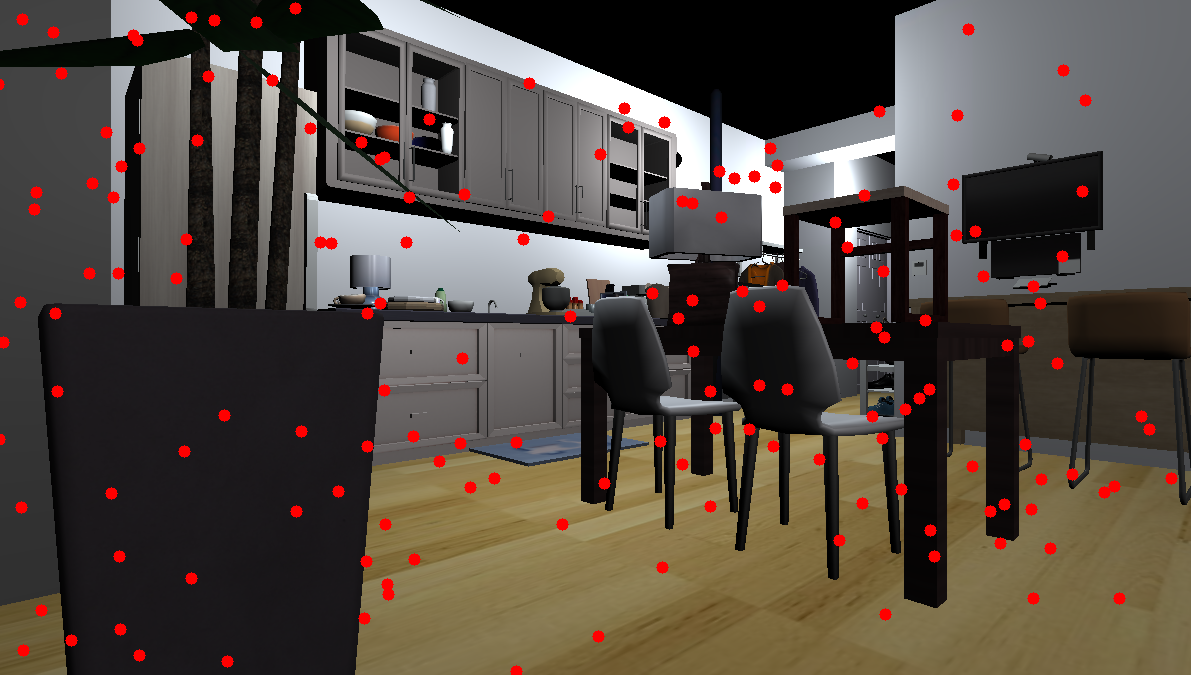}}
    \caption{Active pixel sampling. The irregularity ${\xi}_{b}$ of each block is inferred from the depth image $\mathbb{D}$ and the normal rendering image $\mathbb{N}$ to allocate more sample pixels in more complex regions.}
\label{sampling}
\end{figure}

We first perform pixel sampling for the current frame consisting of a camera pose $T_{wc}$ and a depth image $\mathbb{D}$. 
\textcolor{black}{Current pixel sampling methods, such as complete randomness \cite{nice-slam} or loss-based \cite{imap}, overlook the variation in geometric complexity among objects, an essential factor for effective network learning. In response, we have revamped the pixel sampling strategy. A guideline is to focus more on uneven foregrounds than on flat backgrounds}, which is relatively simpler to fit. For this purpose, the depth image $\mathbb{D}$ is divided into $8*8$ blocks, where the irregularities are evaluated. 
Depth images themselves are reliable candidates for detecting irregularities since they can clearly outline objects and instances in an environment. But the depth information is powerless for distinguishing objects with complex structures (See Fig. \ref{sampling_11} golden circle). So we introduce normal rendering image $\mathbb{N}$, where the value of each pixel is the dot product of the viewing direction vector and the normal direction vector. Comprehensively, the irregularity ${\xi}_{b}$ of each block $b$ is expressed as a weighted sum of the variances of the depth and the normal rendering:
\begin{equation}
\begin{aligned}
    \label{irregularity}
    {{\xi}_{b}}={{\lambda }_{d}} \operatorname{Var}(\mathbb{D}[b])+{{\lambda }_{n}} \operatorname{Var}(\mathbb{N}[b])
\end{aligned}
\end{equation}
\textcolor{black}{where ${\lambda }_{d}$ and ${\lambda }_{n}$ represent the weights.}

Taking the normalized irregularity as the probability of block $b$ being sampled, \textcolor{black}{the number ${M}_{b}$ of sampling points in $b$ can be obtained in (\ref{sampling_piexl})}, where a total of $M$ pixels are sampled in each frame. Fig. \ref{sampling_22} shows a demonstration of active pixel sampling. As expected, more sampled pixels are distributed near skeletal objects such as tables and chairs, while fewer are located on smooth surfaces such as walls and floors. 
\begin{equation}
\begin{aligned}
    \label{sampling_piexl}
    {{M}_{b}}=M\frac{{{\xi}_{b}}}{\sum\nolimits_{b'}{{{\xi}_{b'}}}}
\end{aligned}
\end{equation}

\subsubsection{Point Sampling}
Given the camera internal parameter matrix $K$ and the camera pose $T$, each sampled pixel $[u,v]$ is associated with a ray with the viewing direction $\textbf{r}_{[u,v]}=T_{wc}{{K}^{-1}}[u,v]$ in the world coordinate. \textcolor{black}{In reference to \cite{isdf}, for each ray, we sample $N_f+N_s+1$ points in its viewing direction.} The depth values $l_i$ of these points are composed of $N_f$ stratified samples in range $[l_{min},\mathbb{D}[u,v]+\delta]$ \textcolor{black}{($\delta$ denotes the maximum allowed sampling distance behind the surface)}, $N_s$ near-surface samples satisfying the Gaussian distribution $\mathcal{N}[\mathbb{D}[u,v],\sigma^2]$ \textcolor{black}{($\sigma$ denotes the sampling distance near the surface)} and the surface depth itself $\mathbb{D}[u,v]$ to represent the full observation space and provide more surface supervision. Therefore, the final sampled points are derived as:
\begin{equation}
\begin{aligned}
    \label{sampling_point}
    \textbf{p}_i = l_i\textbf{r}_{[u,v]}
\end{aligned}
\end{equation}

\subsection{Local Updating}
\subsubsection{Capture Distance}
\textcolor{black}{Capturing signed distance for these sampled points $\textbf{p}$ described above is a prerequisite for creating an ESDF.} Unlike Voxblox \cite{voxblox}, in order to more accurately approximate distances, instead of using the depth difference of the rays $\operatorname{sgn}(\mathbb{D}[u,v]-l_i)|\mathbb{D}[u; v]-l_i|$, we directly compute the distance to the nearest surface point:
\begin{equation}
\begin{aligned}
    \label{point_sdf}
    d(\textbf{p})=\operatorname{sgn}(\mathbb{D}[u,v]-l_i)\cdot \left|\textbf{p}-\underset{\textbf{x}\in \mathcal{X}}{\mathop{\arg \min |\textbf{p}-\textbf{x}|}}\,\right|
\end{aligned}
\end{equation}
where $\mathcal{X}$ is the set of surface points sampled in the current frame (i.e. $\textbf{x}=\mathbb{D}[u,v]\textbf{r}_{[u,v]}$).

\begin{figure}[!t]\centering
    \includegraphics[width=8.8cm]{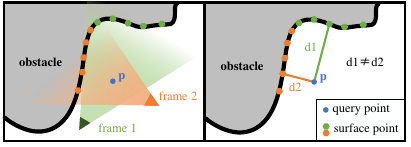}
    \caption{
    There are differences in the approximate distances of the same query point in different frames. If these conflicting data are fed directly into the network, the performance of the implicit map will be affected.}
\label{diffsdf}
\end{figure}

Using point $\textbf{p}$ and distance $d$ directly as training data may lead to degraded network learning performance because $d$ only represents a realistic approximation for the given current frame, while there may be large differences in other frames, as shown in Fig. \ref{diffsdf}.  To alleviate data conflicts, we utilize grids to fuse approximate distance values of sampled points with similar coordinates throughout the observation period.

\subsubsection{SDF Local Fusion}
The perceived space is discretized into axis-aligned grids for local fusion of the SDF. The update equations \eqref{distance_update} and \eqref{weight_update} are based on the existing distance $D$ and weight $W$ of grid $\textbf{v}$, and the distance $d$ and weight $\omega$ of sampled point \textbf{p} falling into grid $\textbf{v}$. \textcolor{black}{We set a maximum value ${W}_{\max }$ for the weight $W$ to ensure the dynamic updating capability of the grid, even if it is observed multiple times by history data.} Distances are generally overestimated, so the heuristic weight function as \eqref{weight} is designed, \textcolor{black}{where $\varepsilon$ and ${{w}_{\min }}$ are empirical parameters.} The distances of points sampled from different frames but falling into the same grid would be merged and associated with the grid center. Over time, ${D}$ gradually approaches the ground truth by fusing observations from different frames. 
\begin{equation}
\begin{aligned}
    \label{weight}
    \omega(\textbf{p})=\max \left( \exp (-\varepsilon \cdot |d(\textbf{p})|),{{w}_{\min }} \right) 
\end{aligned}
\end{equation}
\begin{equation}
\begin{aligned}
    \label{distance_update}
    {{D}_{t}}(\textbf{v},\textbf{p})=\frac{{{W}_{t-1}}(\textbf{v}){{D}_{t-1}}(\textbf{v})+\omega (\textbf{p})d(\textbf{p})}{{{W}_{t-1}}(\textbf{v})+\omega (\textbf{p})}
\end{aligned}
\end{equation}
\begin{equation}
\begin{aligned}
    \label{weight_update}
    {{W}_{t}}(\textbf{v},\textbf{p})=\min \left( {{W}_{t-1}}(\textbf{v})+\omega (\textbf{p})),{{W}_{\max }} \right)
\end{aligned}
\end{equation}

Additionally, we update the gradient $\textbf{G}$ towards the nearest obstacle for grids, which is useful for subsequent network training. In our formulation, the gradient $\textbf{g}$ of each sampled point is the direction to the nearest surface point. For surface points, we replace $\textbf{g}$ with surface normals computed from the depth image $\mathbb{D}$. The gradient is updated in the same way as distance:
\begin{equation}
\begin{aligned}
    \label{point_G}
    {{\textbf{G}}_{t}}(\textbf{v},\textbf{p})=\frac{{{W}_{t-1}}(\textbf{v}){{\textbf{G}}_{t-1}}(\textbf{v})+\omega (\textbf{p})\textbf{g}(\textbf{p})}{{{W}_{t}}(\textbf{v})+\omega (\textbf{p})}
\end{aligned}
\end{equation}

\subsection{Global Learning}

\subsubsection{Select Train Grids}
The ultimate goal of LGSDF is to train the implicit ESDF neural network ${f}_{\theta}$ so that it has sufficient ability to represent the current scene. In the previous modules, we transformed raw depth observations into a dynamically updated discrete explicit ESDF map composed of grids $\textbf{v}$. The grids with weights $W>0$ are the updated grids from which we select the training data batches.

\begin{figure}[!t]\centering
    \includegraphics[width=8.8cm]{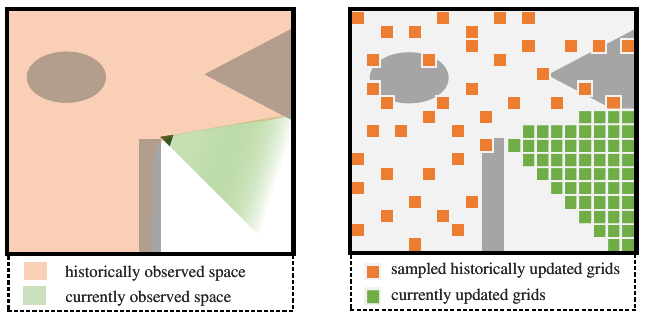}
    \caption{An example of selecting training grids. Left: The current awareness of the scene. Right: The selected grids for training. 
    The Currently updated grids and the sampled historically updated grids guarantee good plasticity and stability of the network, respectively.} 
\label{sele_grid}
\end{figure}

The selected grids consist of two parts as shown in Fig. \ref{sele_grid}. 
The first part is all the currently updated grids, which ensure good plasticity of the network for faster dynamic adjustment. These grids are either revised or unfamiliar, so they deserve to be studied intensively.
The second part comes from historically updated grids, which guarantee good stability of the network to preserve previous knowledge. We randomly sample historically updated grids to ensure that each of them has an equal chance of being selected, and that the selected grids are dispersed as widely as possible across the historical observation space.
The iterative learning of the historically updated grids not only avoids catastrophic forgetting in continual learning, but also stabilizes the overall structure of the scene.

\subsubsection{Loss Function}
For each input frame, we perform $N_I$ iterations and design several ESDF generic losses for self-supervised learning.

(i) SDF Loss: As a direct manifestation of mapping quality, SDF loss is the most critical item. We distinguish between free-space and near-surface grids. Despite approaching the true distances through local fusion, the existing distances $D$ of free-space grids remain strict upper bounds. Therefore, the SDF loss for free-space grids performs a penalty when their predicted distances exceed bounds or become negative:
\begin{equation}
\begin{aligned}
    \label{sdf_loss_free}
    {{\mathcal{L}}_{\text{f}\_\text{s}}}({{f}_{\theta }}(\textbf{v}),D)=\max (0,{{e}^{-\beta {{f}_{\theta }}(\textbf{v})}}-1,{{f}_{\theta }}(\textbf{v})-D)
\end{aligned}
\end{equation}
\textcolor{black}{where $\beta$ is a function parameter.} For those close to the surface, their distances are more reliable, so we enforce stricter supervision directly through $L1$ loss:
\begin{equation}
\begin{aligned}
    \label{sdf_loss_near}
    {{\mathcal{L}}_{\text{n\_s}}}({{f}_{\theta }}(\textbf{v}),D)=\left| {{f}_{\theta }}(\textbf{v})-D \right|
\end{aligned}
\end{equation}
Taking $T$ as the truncation distance of free-space and near-surface, \textcolor{black}{and ${\lambda }_{\text{n\_s}}$ as the free-space loss weights,} the full SDF loss is:
\begin{equation}
\begin{aligned}
    \label{sdf_loss}
    {{\mathcal{L}}_{\text{sdf}}}({{f}_{\theta }}(\textbf{v}),D)=\begin{dcases}
    {{\lambda }_{\text{n\_s}}} {{\mathcal{L}}_{\text{n\_s}}} & \left| D \right|<T \\
     {{\mathcal{L}}_{\text{f\_s}}} & \text{otherwise}
    \end{dcases}
\end{aligned}
\end{equation}

(ii) Gradient Loss: In ESDF, the gradient indicates the direction to the nearest obstacle, which can be used to optimize the signed distance values. The gradient of the SDF predicted by the network can be efficiently computed by automatic differentiation. Therefore, we introduce the cosine distance of two vectors as the gradient loss:
\begin{equation}
\begin{aligned}
    \label{grad_loss}
    {{\mathcal{L}}_{\text{grad}}}({{\nabla }_{\textbf{v}}}{{f}_{\theta }}(\textbf{v}),\textbf{G})=1-\frac{{{\nabla }_{\textbf{v}}}{{f}_{\theta }}(\textbf{v})\cdot \textbf{G}}{\left\| {{\nabla }_{\textbf{v}}}{{f}_{\theta }}(\textbf{v}) \right\|\left\| \textbf{G} \right\|}
\end{aligned}
\end{equation}

(iii) Eikonal Loss: Full SDF is differentiable almost everywhere, and its gradient satisfies the eikonal equation. The regularization of the Eikonal equation is similar to the wavefront algorithm, which serves to propagate the field from near the surface to free space. Since gradient discontinuities that do not satisfy the Eikonal equation are more common near the surface, we perform the penalty only in free space:
\begin{equation}
\begin{aligned}
    \label{eikonal_loss}
      {{\mathcal{L}}_{\text{eik}}}({{\nabla }_{\textbf{g}}}{{f}_{\theta }}(\textbf{v}))=
    \begin{dcases}
  \left| \left\| {{\nabla }_{\textbf{g}}}{{f}_{\theta }}(\textbf{g}) \right\|-1 \right| & \left| D \right| \ge T \\
    0 & \text{otherwise}
    \end{dcases}
\end{aligned}
\end{equation}

\textcolor{black}{In each iteration, the network parameters $\theta$ are optimized minimizing the weighted sum of the three losses:}
\begin{equation}
\begin{aligned}
    \label{all_loss}
	l(\theta )={{\mathcal{L}}_{\text{sdf}}}+{{\lambda }_{\text{grad}}}{{\mathcal{L}}_{\text{grad}}}+{{\lambda }_{\text{eik}}}{{\mathcal{L}}_{\text{eik}}}
\end{aligned}
\end{equation}
\textcolor{black}{where ${\lambda }_{\text{grad}}$ and ${\lambda}_{\text{eik}}$ are the weights of the gradient loss and eikonal loss, respectively.}

\section{Experimental Results} \label{ER}
In this section, the performance of the proposed LGSDF algorithm is validated through experiments performed on multiple scenes.

\subsection{Experimental Setup}

\textbf{Evaluation Scenes:} \textcolor{black}{We utilize the same public scenes as iSDF \cite{isdf}, including two scenes (\textit{apt\_2} and \textit{apt\_3}) from the synthetic dataset \textbf{ReplicaCAD} \cite{habitat}, and six scenes (\textit{scene\_0004}, \textit{scene\_0005}, \textit{scene\_0009}, \textit{scene\_0010}, \textit{scene\_0030} and \textit{scene\_0031}) from the real-world dataset \textbf{ScanNet} \cite{scannet}. The ReplicaCAD dataset generates depth images of dimensions $1200*680$ by simulating a locobot robot equipped with a depth sensor navigating through two distinct apartment configurations. The camera poses in this dataset utilize Ground Truth (GT) poses provided by the emulator. On the other hand, the ScanNet dataset employs the Structure Sensor to capture depth images with dimensions of $640*480$ in the real environment.  The camera poses in ScanNet dataset are obtained using the BundleFusion algorithm \cite{bundlefusion}.} 

\textbf{Comparison Baseline:} \textcolor{black}{For explicit mapping, we chose several classical ESDF mapping methods \textbf{Voxblox} \cite{voxblox}, \textbf{FIESTA} \cite{FIESTA} and \textbf{VDB-EDT} \cite{vdb} as baselines. For implicit mapping, the most similar to our work is \textbf{iSDF} \cite{isdf}, which also strives to build globally continuous ESDF maps using a simple MLP. 
In addition, we have chosen two local implicit mapping methods, \textbf{Go-surf} \footnote{Since the open source code of Go-surf does not provide the option of incremental mapping, we randomly select frames from all the frames according to the original settings.} \cite{go-surf} and \textbf{SHINE-Mapping} \cite{shine} (abbreviated as S.-M. in tables and figures). Similarly, they provide signed distance values for the constructed region in order to construct the obstacle surfaces in the scene.}

\textbf{Implementation Details:} \textcolor{black}{All experiments are carried out on a desktop system with an Intel(R) Xeon(R) Platinum 8336C base 2.30GHz CPU and an NVIDIA Geforce RTX 4090 GPU.}
\textcolor{black}{To ensure fairness in the experiments, all methods simulate practical applications running in real time. Specifically, for all explicit mapping methods \cite{voxblox,FIESTA,vdb}, we convert the dataset from file format to rosbag format for real-time play. For iSDF \cite{isdf}, SHINE-Mapping and our method,  we continuously process the latest frames based on the runtime and the dataset frame rate and stop optimizing when the dataset reaches its end. As for Go-surf \cite{go-surf}, the batch training time is forced to be the same as the data acquisition time to maintain consistency.} 

\textbf{Generalization Note:} 
\textcolor{black}{It is worth noting that all of the above implicit mapping methods are scene-specific, which means that they do not require additional pre-training and can gradually recover the scene during continual learning. This approach can efficiently reconstruct the observed specific scenes without the need to generalize across different environments.}

\textbf{Evaluation Metric:} 
We choose two metrics for quantitative evaluation: SDF error and mesh completeness.
SDF error is the mean absolute difference of the distance of the sampled locations between the predicted ESDF and the ground truth ESDF.
Mesh completeness is the mean absolute distance of densely sampled points from the ground truth mesh to the nearest surface on the predicted mesh. 

\textbf{Experience Parameters:} In all experiments, we set parameters as follows: ${\lambda }_{d}=0.7$, ${\lambda }_{n}=0.3$, $M=200$, $N_f = 22$, $N_s = 5$, $l_{min} = 7cm$, $\delta = 10cm$, $\sigma = 10cm$, $\varepsilon=50$, ${w}_{\min }=0.00001$, ${W}_{\max } = 10$, $\beta = 5$, $T=10cm$, ${\lambda }_{\text{n\_s}}=3.38$, ${\lambda }_{\text{grad}}=0.00018$, ${\lambda }_{\text{eik}}=0.0268$ and $N_I=10$. We use a learning rate of $0.0013$ and a weight decay of $0.012$ for the Adam optimizer.  These parameters are partly based on previous work and partly on experience.

\begin{figure}[!t]\centering
	\includegraphics[width=8.5cm]{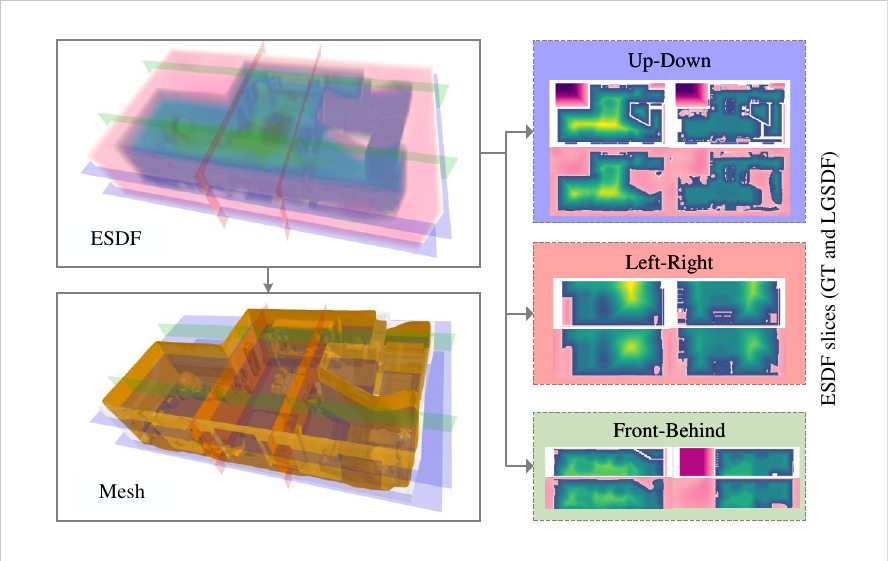}
	\caption{Visualization of implicit maps constructed by LGSDF in \textit{apt\_3}. 3D ESDF indicates that the output of the LGSDF is a true full SDF. } 
	\label{demo} 
\end{figure}

\subsection{ESDF and Mesh}

As mentioned in \ref{AO}, we visualize the implicit map constructed by LGSDF in two forms, which are discrete ESDF with pre-defined resolution and zero-level set mesh. Taking \textit{apt\_3} as an example, Fig. \ref{demo} shows both ESDF and mesh. The 3D ESDF indicates that the output of the LGSDF is a true full SDF, providing the predicted signed distance at any continuous location. We extract six 2D ESDF slices and compare them with ground truth. LGSDF not only outputs almost exactly correct results in the visible region but also makes its inferences in the invisible region. The mesh constructed with the marching cubes \cite{marching_cubes} almost fits the real simulation scene.

\begin{figure*}[!t]\centering
	\includegraphics[width=17.5cm]{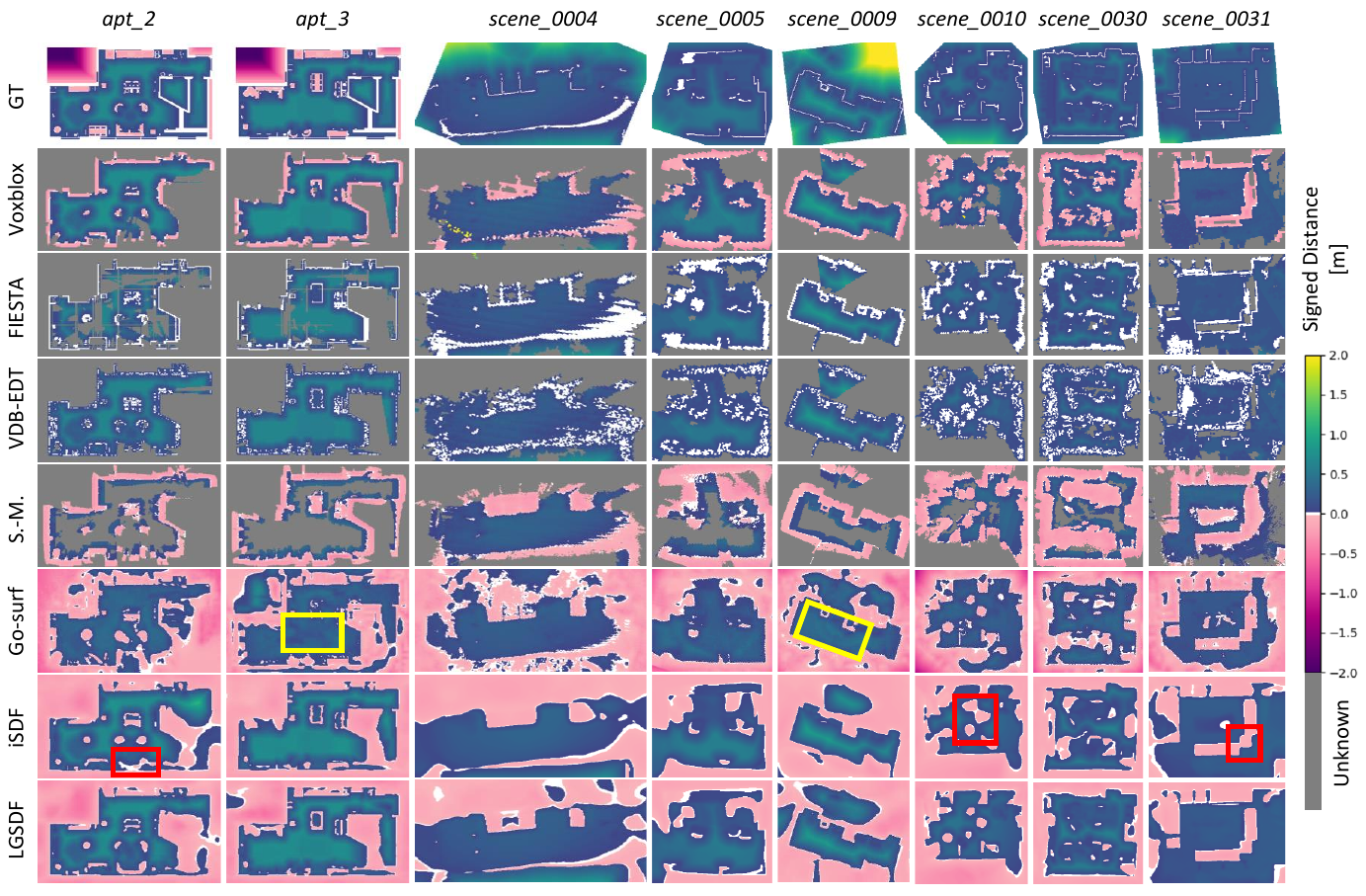}
	\caption{\textcolor{black}{2D ESDF slices for different methods. As explicit mapping methods, Voxblox, FIESTA and VDB-EDT cannot reasonably predict the SDF of invisible regions. Octree-based SHINE-Mapping can only provide truncated SDFs. Go-surf is unable to produce accurately signed distance values in free space (yellow boxes). iSDF maps are cluttered (red boxes). LGSDF always maps the results closest to the ground truth.}}
	\label{slice} 
\end{figure*}

\begin{figure*}[!t]\centering
	\includegraphics[width=17.5cm]{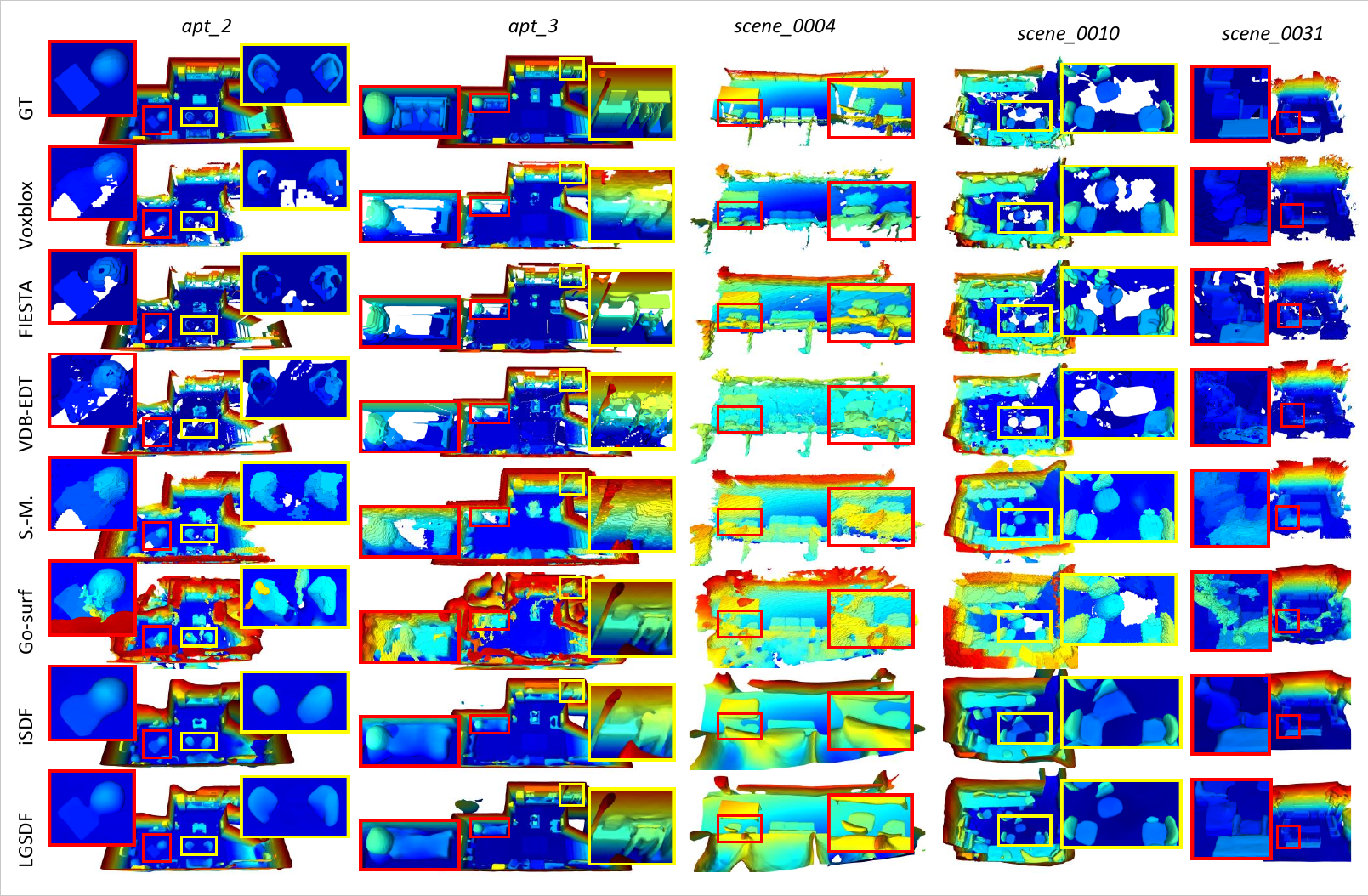}
	\caption{\textcolor{black}{Zero-level set meshes for different methods. Voxblox, FIESTA, VDB-EDT and SHINE-Mapping construct incomplete and noisy meshes. Meshes of Go-surf have large artifacts and are not smooth and regular. Objects in iSDF are distorted. Different from these, LGSDF recovers smooth continuous and accurate meshes.}}
	\label{mesh} 
\end{figure*}

\subsection{\textcolor{black}{ESDF Slices Qualitative results}}

To compare the reconstructed SDFs by different methods, Fig. \ref{slice} presents 2D ESDF slices at a constant height for all scenes. 
\textcolor{black}{As explicit mapping methods, Voxblox, FIESTA, and VDB-EDT are limited to mapping only the visible region. While they propagate signed distance values from the obstacle's surface to the surrounding free area, they are unable to address the invisible parts due to perceptual constraints. Consequently, the query for distance values remains invalid in numerous regions, as depicted in gray.}

\textcolor{black}{The 2D ESDF slices obtained through SHINE-Mapping yield similar outcomes. This is attributed to SHINE-Mapping utilizing an octree as its underlying representation, storing high-dimensional features obtained through encoding. Although this information can be utilized to derive signed distance values using a simple decoder, SHINE-Mapping tends to assign nodes only near the obstacle's surface, resulting in an incomplete ESDF. In contrast, Go-surf employs multi-resolution voxel hashes across the entire scene space, enabling the generation of distance values for arbitrary locations. 
Nevertheless, the incorporation of a 2D rendering loss in Go-surf leads to a reduction in SDF accuracy as it disperses the training focus. Furthermore, the absence of gradient loss causes the predicted signed distance values in the free space to deviate significantly, as evident in the yellow boxes in Fig. \ref{slice}.}

\textcolor{black}{As global implicit mapping methods, both iSDF and LGSDF show more competitive results, with both generating smoother and more continuous ESDFs. However, as shown in the red boxes in Fig. \ref{slice}, the signed distance of iSDF is disordered.} An important reason for this phenomenon is that the training data used by iSDF are conflicting, and replay aggravates the consequences. In contrast, the data associated with the grids in LGSDF is always the optimal solution at the current moment. Furthermore, the uniform and regular grid arrangement avoids forgetting any visible region. As a result, LGSDF always maps the results closest to the ground truth.

\subsection{\textcolor{black}{Meshes Qualitative Results}}

The zero-level set meshes of representative scenes are shown in Fig. \ref{mesh}. Their effects more clearly reflect the above conclusions. 
\textcolor{black}{All the explicit mapping methods and SHINE-Mapping operate discretely, posing challenges in reconstructing surfaces that are continuous and smooth. Their meshes often exhibit loopholes, indicating the lack of available data at specific locations. 
Notably, SHINE-Mapping's imperfections in this regard tend to be slightly smaller than those observed in other explicit mapping methods. This discrepancy arises from SHINE-Mapping's utilization of trilinear interpolation across various octree levels to ascertain the attributes of query point locations, thereby preserving continuity within the nodes.}

\textcolor{black}{While Go-surf constructs meshes continuously, they lack regularity and smoothness. The multi-resolution voxel feature employed by Go-surf keeps the voxels essentially discrete, resulting in a rough surface for the object. Simultaneously, this method tends to fabricate artifacts in unobserved areas.}

\textcolor{black}{Diverging from the local implicit mapping strategy, both iSDF and LGSDF adopt a single MLP as the representational form, which enables information about unobserved regions to be inferred from existing training data. Moreover, since the output of the network is continuous in 3D, the generated meshes exhibit global smoothness. However, it's worth noting that iSDF meshes display distortions and sticking issues, diminishing the practicality of the generated maps. In contrast, LGSDF excels in providing more precise boundaries for objects.}

\begin{figure*}[!t]\centering
	\includegraphics[width=18.3cm]{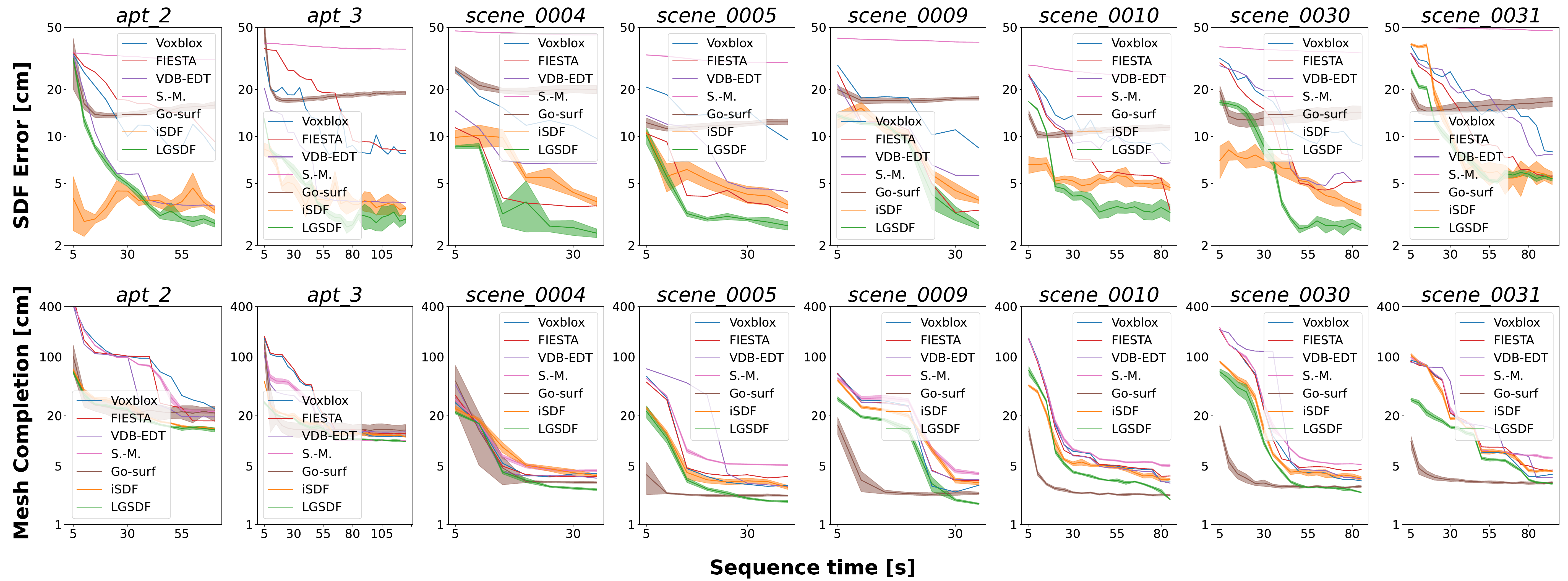}
	\caption{\textcolor{black}{Continuous dynamic changes during the incremental mapping process for different methods along two evaluation metrics: SDF error and mesh completion. LGSDF achieves the best results in every scene.}}
	\label{plot} 
\end{figure*}

\begin{table*}[!t]
\scriptsize
\renewcommand{\arraystretch}{1.2}
\centering
\caption{\textcolor{black}{The Final Evaluation Results Of Different Scenes}}
\label{all_results}
\resizebox{\textwidth}{!}{\begin{tabular}{p{1.4cm}<{\centering}p{1.7cm}<{\centering}p{1.2cm}<{\centering}p{1.2cm}<{\centering}p{1.2cm}<{\centering}p{1.2cm}<{\centering}p{1.2cm}<{\centering}p{1.2cm}<{\centering}p{1.2cm}<{\centering}p{1.2cm}<{\centering}}
\hline
\textbf{Metrics} &
  \textbf{Methods} &
  \textit{apt\_2} &
  \textit{apt\_3} &
  \textit{scene\_0004} &
  \textit{scene\_0005} &
  \textit{scene\_0009} &
  \textit{scene\_0010} &
  \textit{scene\_0030} &
  \textit{scene\_0031} \\ \hline
 & \textbf{Voxblox} & 08.09 & 07.74 & 09.69 & 09.50 & 08.43 & 08.09 & 08.74 & 07.97 \\
 & \textbf{FIESTA}  & 09.31 & 08.15 & 03.59 & 03.23 & 03.37 & 03.38 & 05.12 & 05.58 \\
 & \textbf{VDB-EDT} & 03.58 & 03.79 & 06.74 & 04.44 & 05.62 & 06.77 & 05.24 & 07.64 \\
 & \textbf{S.-M.}   & 31.03 & 36.28 & 44.74 & 29.71 & 40.19 & 24.02 & 34.31 & 47.78 \\
 & \textbf{Go-surf} & 15.89 & 19.00 & 19.99 & 12.40 & 17.56 & 11.37 & 14.30 & 16.67 \\
 & \textbf{iSDF}    & 03.38 & 03.46 & 03.81 & 03.65 & 03.90 & 04.70 & 03.38 & 05.42 \\
  \multirow{-7}{*}{\textbf{\begin{tabular}[c]{@{}c@{}}SDF\\ Error\\ {[}cm{]} \\ $\downarrow$\end{tabular}}} & \textbf{LGSDF}   & \textbf{02.77} & \textbf{02.95} & \textbf{02.40} & \textbf{02.68} & \textbf{02.69} & \textbf{03.26} & \textbf{02.59} & \textbf{05.35} \\ \hline
& \textbf{Voxblox} & 24.26 & 11.33 & 04.06 & 02.96 & 02.96 & 03.24 & 03.34 & 03.99 \\
 & \textbf{FIESTA}  & 17.29 & 12.27 & 03.73 & 03.75 & 03.38 & 03.81 & 04.54 & 04.49 \\
 & \textbf{VDB-EDT} & 19.31 & 12.84 & 03.59 & 02.86 & 03.45 & 03.13 & 03.56 & 03.65 \\
 & \textbf{S.-M.}   & 22.32 & 10.05 & 04.41 & 05.13 & 04.08 & 05.11 & 05.24 & 06.25 \\
 & \textbf{Go-surf} & 21.51 & 13.47 & 03.19 & 02.22 & 02.37 & 02.26 & 02.86 & 03.19 \\
 & \textbf{iSDF}    & 13.91 & 11.15 & 03.85 & 02.79 & 03.17 & 03.48 & 03.59 & 04.33 \\
\multirow{-7}{*}{\textbf{\begin{tabular}[c]{@{}c@{}}Mesh\\ Completion\\ {[}cm{]} \\ $\downarrow$\end{tabular}}}
& \textbf{LGSDF}   & \textbf{13.54} & \textbf{09.82} & \textbf{02.62} & \textbf{01.90} & \textbf{01.77} & \textbf{02.00} & \textbf{02.43} & \textbf{03.09} \\ \hline
\end{tabular}
}
\end{table*}

\subsection{Quantitative Results}

Taking $5s$ as the evaluation period, Fig. \ref{plot} shows the continuous dynamic changes of the two metrics during the incremental mapping process. \textcolor{black}{For the implicit mapping methods, the effect of learning varies each time due to sampling uncertainty, so we repeat their experiments $5$ times. The signed distance of the unknown region defaults to $0$.} As the observations gradually cover the entire scene, the performance of all methods on both metrics is improved overall. \textcolor{black}{For a clearer comparison, we report the final evaluation results in Tab. \ref{all_results}.}

LGSDF achieves the best results in all scenes. 
\textcolor{black}{
SHINE-Mapping exhibits the highest SDF error primarily because it can only capture values in the immediate vicinity of the obstacle's surface. In the realm of SDF errors, Go-surf secures the second-to-last position, primarily attributed to a notable bias in the free region. Remarkably, this bias does not negatively impact its mesh reconstruction quality, as evidenced by the mesh completion.
Comparatively, the other methods do better in terms of SDF quality and vary in terms of mesh completion. As the main baseline, iSDF lags behind LGSDF in both metrics, aligning with the observed qualitative results above.
}


\begin{figure}[!t]\centering
    \subfigure[full LGSDF]{\label{ablation_11}
    \includegraphics[width=2.75cm]{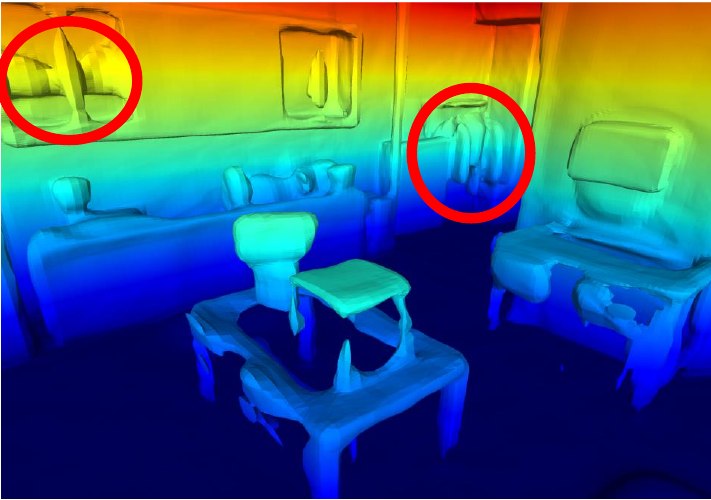}}
    \subfigure[random sampling ]{\label{ablation_12}
    \includegraphics[width=2.75cm]{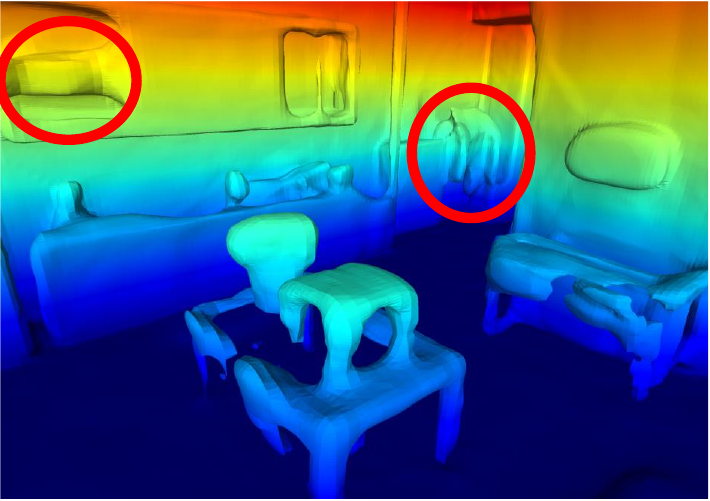}}
    \subfigure[w/o current grids ]{\label{ablation_21}
    \includegraphics[width=2.75cm]{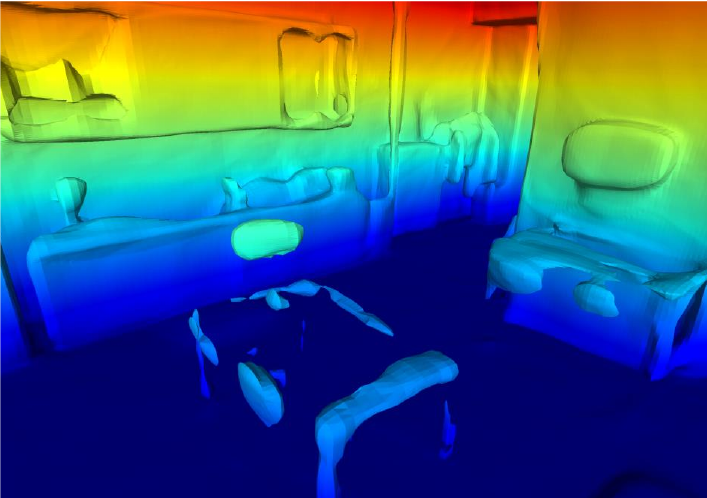}}
    \vfill
    \subfigure[w/o history grids]{\label{ablation_22}
    \includegraphics[width=2.75cm]{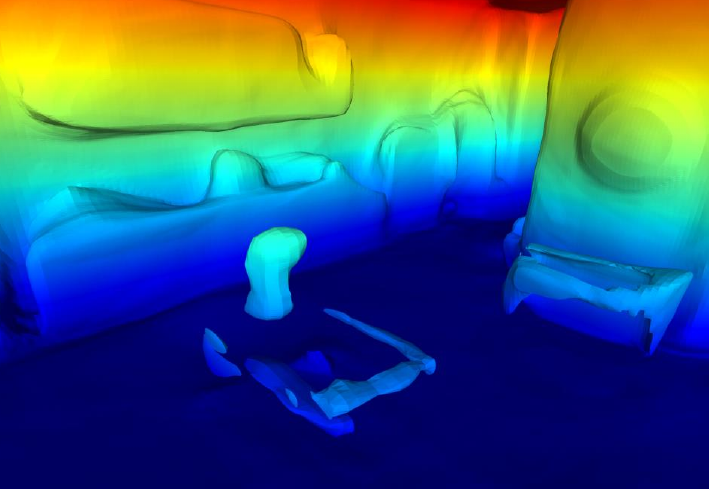}}
    \subfigure[w/o grids w replay]{\label{ablation_31}
    \includegraphics[width=2.75cm]{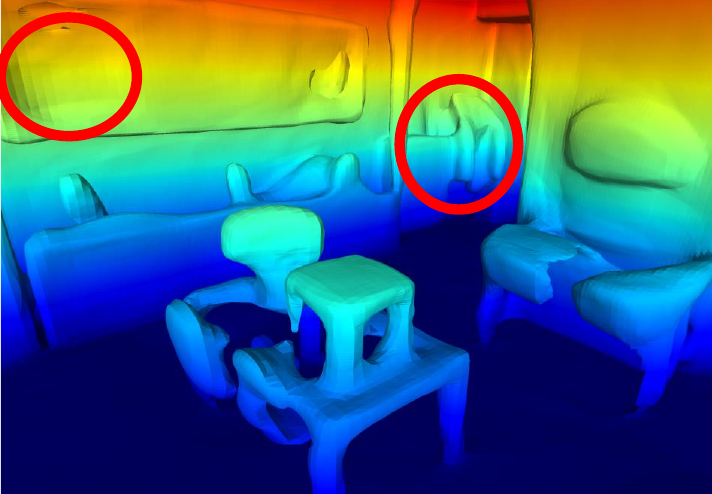}}
    \subfigure[w/o grids w/o replay]{\label{ablation_32}
    \includegraphics[width=2.75cm]{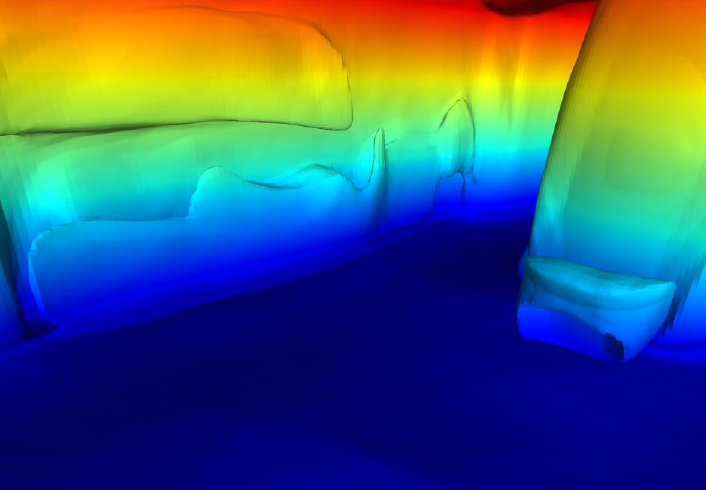}}
    \caption{\textcolor{black}{The ablation experiments in secne \textit{apt\_3}. Both the irregularity-based pixel sampling strategy and the grid-based local fusion strategy contribute to the map quality.}}
\label{ablation}
\end{figure}

\begin{table}[!t]
\scriptsize
\renewcommand{\arraystretch}{1.1}
\centering
\caption{\textcolor{black}{Quantitative Results of Ablation Experiments}}
\label{ablation_table}
\resizebox{\columnwidth}{!}{
\begin{tabular}{ccc}
\hline
 & \textbf{SDF Error [cm]} & \textbf{Mesh Completion [cm]}\\ \hline
full LGSDF & \textbf{02.95} & \textbf{09.82} \\
random sampling & 03.29 & 10.46 \\
w/o current grids & 03.11 & 10.68 \\
w/o history grids & 08.10 & 15.16 \\
w/o grids w replay & 03.13 & 10.71 \\
w/o grids w/o replay & 24.88 & 17.39 \\ \hline
\end{tabular}
}
\end{table}

\subsection{\textcolor{black}{Ablation Study}}

\textcolor{black}{To evaluate the role of each proposed module, we performed ablation experiments in the \textit{apt\_3} scene and the results are shown in Fig. \ref{ablation} and the metrics are shown in Tab. \ref{ablation_table}.}

\subsubsection{\textcolor{black}{Impact of irregularity-based pixel sampling strategy}}
\textcolor{black}{We assess the influence of the irregularity-based pixel sampling strategy. As depicted by the highlighted red circles in Fig. \ref{ablation_12}, substituting this strategy with random pixel sampling results in a decrease in the quality of certain map details. This is attributed to the fact that the random pixel sampling strategy does not allocate a greater number of pixels to these regions. Instead, it treats them on par with walls or floors, which are comparatively easier to fit. In other words, it lacks enough observations to reconstruct the details.}

\subsubsection{\textcolor{black}{Impact of grids}}
\textcolor{black}{To investigate the impact of grids, we conducte four specific experiments. Excluding the currently observed grids during training introduces the risk of omitting specific objects (Fig. \ref{ablation_21}). The omission of historically updated grids in training significantly increases the likelihood of optimization failure (Fig. \ref{ablation_22}), primarily due to the network's susceptibility to catastrophic forgetting during continual learning. Moreover, the absence of a grid-based local fusion strategy results in conflicts within the training data, leading to a degradation of map detail accuracy (Fig. \ref{ablation_31}) even with the introduction of history frame replay. Furthermore, if the replay mechanism is not utilized, the map undergoes additional deterioration (Fig. \ref{ablation_32}).}

\begin{figure}[!t]\centering
	\includegraphics[width=7.3cm]{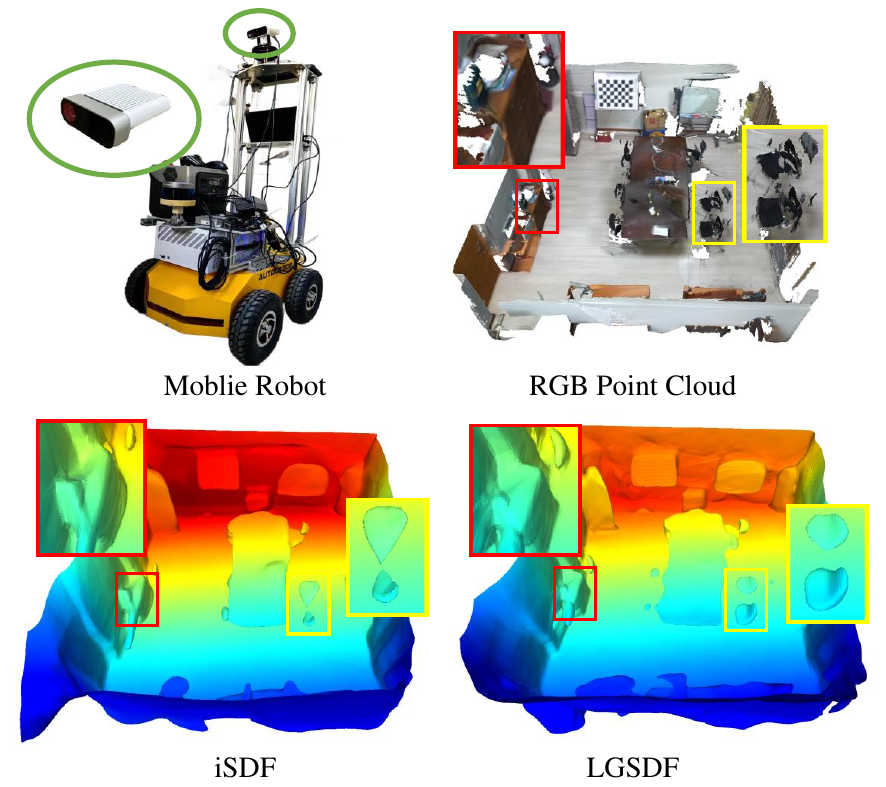}
	\caption{\textcolor{black}{Results of experiments with self-capture datasets. LGSDF shows more competitiveness than iSDF in reconstructing object details.}}
	\label{conference} 
\end{figure}

\subsection{\textcolor{black}{Self-capturing Dataset Experiment}}

\textcolor{black}{To assess the real-world effectiveness of LGSDF, we conducted experiments within a self-captured \textit{conference\_room} scene using an Autolabor robot equipped with an Azure Kinect DK depth camera, which provides depth images of size $1280*720$. The camera poses are obtained using the Open3D open-source framework. In Fig. \ref{conference}, we present the robot platform and the comparative results of LGSDF against the primary baseline iSDF. 
Despite encountering numerous blind spots in the field of view, both iSDF and LGSDF managed to generate comprehensive and smooth ground maps, leveraging the predictive capabilities of the network.  A closer examination of the map details reveals that the quality of the map produced by LGSDF surpasses that of iSDF. Noteworthy improvements are evident in specific areas, such as the detailed depiction of the left cabinet upon zooming in and the accurate representation of the two chairs on the right.
This observation suggests that LGSDF's approach, incorporating local updating to complement global learning, plays a pivotal role in the continuous and accurate construction of implicit maps.
}

\begin{table}[!t]
\tiny
\renewcommand{\arraystretch}{1.1}
\centering
\caption{\textcolor{black}{Storage and GPU Memory Comparison in Scene\textit{scene\_0005}}}
\label{TM_table}
\resizebox{\columnwidth}{!}{
\begin{tabular}
{p{1.3cm}<{\centering}p{1.5cm}<{\centering}p{1.5cm}<{\centering}}
\hline
 & \textbf{Storage} & \textbf{GPU mempry} \\ \hline
\textbf{Voxblox} & 2 MB & \textbf{-} \\
\textbf{FIESTA} & 2 MB & \textbf{-} \\
\textbf{VDE-EDT} & 2 MB & \textbf{-} \\
\textbf{S.-M.} & 160 MB & 4.5 GB \\
\textbf{Go-surf} & 340 MB & 5.5 GB \\
\textbf{iSDF} & \textbf{1 MB} & 2.8 GB \\
\textbf{LGSDF} & \textbf{1 MB} & \textbf{2.4 GB} \\ \hline
\end{tabular}
}
\end{table}

\subsection{\textcolor{black}{Timings and memory}}

\textcolor{black}{In many application scenarios, real-time constraints and limited memory are common challenges. To assess the computational overhead of various algorithms, we use the scene \textit{scene\_0005} as an example and obtain Tab. \ref{TM_table}.}

\textcolor{black}{When comparing storage, iSDF and LGSDF demonstrate a significant advantage, requiring only $1$ MB for the network weights of the MLP. In contrast, all explicit mapping methods necessitate approximately $2$ MB to store voxel grids. SHINE-Mapping and Go-surf demand $160$ MB and $340$ MB, respectively, for storing a large number of high-dimensional features and corresponding decoders. 
Regarding GPU memory consumption, SHINE-Mapping and Go-surf utilize $5.5$ GB and $4.5$ GB, respectively, in contrast to $2.8$ GB and $2.4$ GB for iSDF and LGSDF. The reason LGSDF consumes less memory than iSDF is attributed to the introduction of local updating, eliminating the need to store historical keyframes for replay.}

\textcolor{black}{In experiments, for fair comparison, all methods take an equal amount of runtime ($38.67$ s for scene \textit{scene\_0005}).  In this process, LGSDF completes each iteration in about $20$ ms (i.e., $0.2$ s per frame, for $N_i=10$), proving its suitability for real-world applications.}

\section{Conclusion} \label{Conclusion}

This paper proposes a novel online continual learning algorithm LGSDF for building ESDF maps. 
First, we actively sample 3D points, which give more attention to irregular regions. Then, the approximate distances of different frames are fused into discrete grids, which alleviates data conflicts and improves local accuracy. Finally, we use these grids to train a neural network in a self-supervised manner, resulting in the generation of the desired implicit map. The verification results in multiple public scenes show that LGSDF can construct the most accurate SDF and zero-level set meshes. 

\textcolor{black}{However, the LGSDF still has some limitations. Like most implicit mapping methods, LGSDF is more suitable for scenes with a limited scope and may struggle to handle dynamic objects effectively. 
In the future, we have plans to extend the capabilities of LGSDF to encompass large-scale dynamic outdoor scenes.}

\bibliographystyle{Bibliography/IEEEtran}
\bibliography{Bibliography/TVCG}

\newpage

\section*{Biography Section}

\begin{IEEEbiography}[{\includegraphics[width=1in,height=1.25in,clip,keepaspectratio]{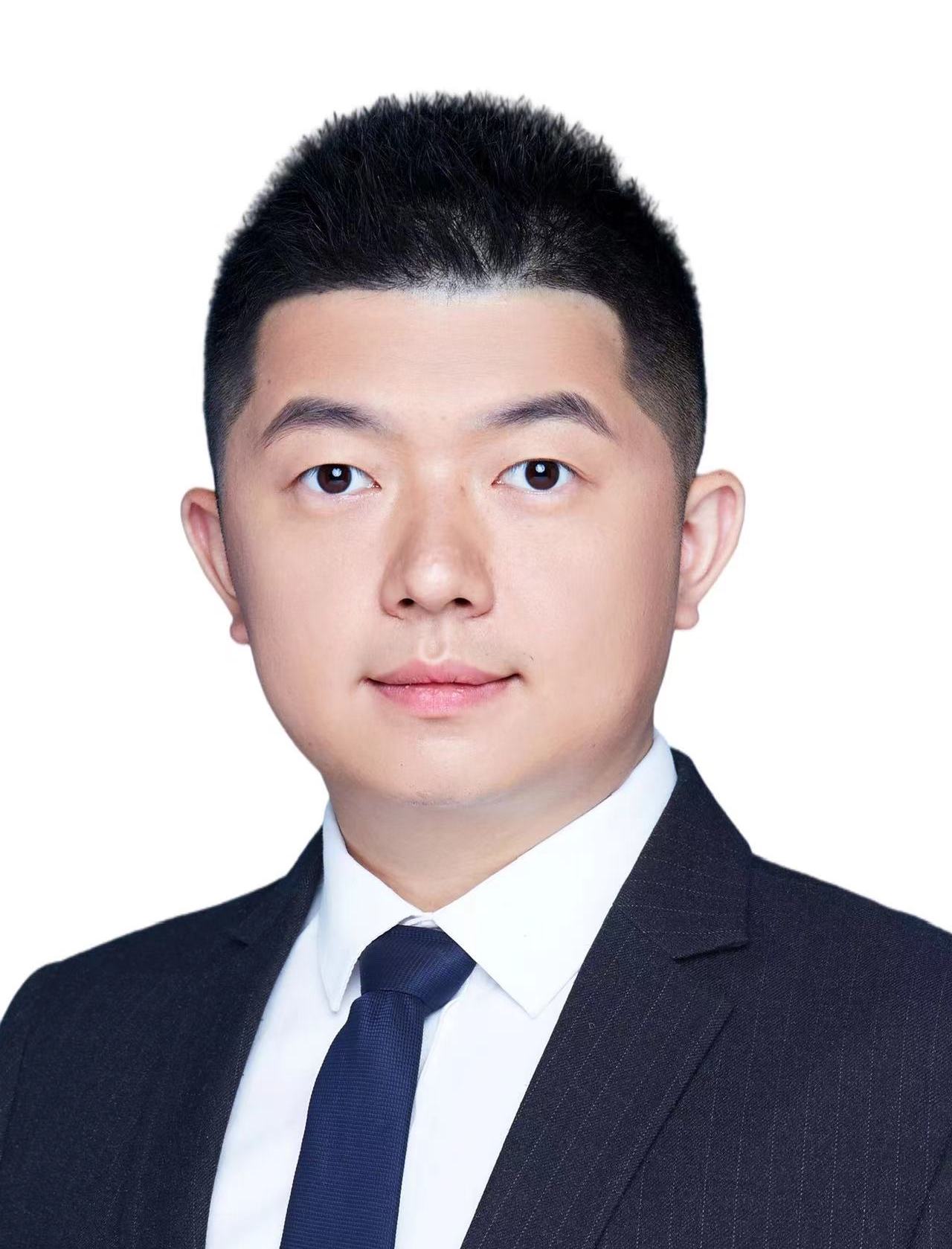}}]{Yufeng Yue} received the B.Eng. degree in automation from Beijing Institute of Technology, Beijing, China, in 2014, the Ph.D. degree from Nanyang Technological University, Singapore, in 2019.  He is currently a Professor with School of Automation, Beijing Institute of Technology. He has published a book in Springer, and over 40 journal/co   nference papers, including TMECH, TIE, TCST, TMM, ICRA and IROS. His research interests include perception, mapping and navigation for collaborative robots. 
 \end{IEEEbiography}

\begin{IEEEbiography}[{\includegraphics[width=1in,height=1.25in,clip,keepaspectratio]{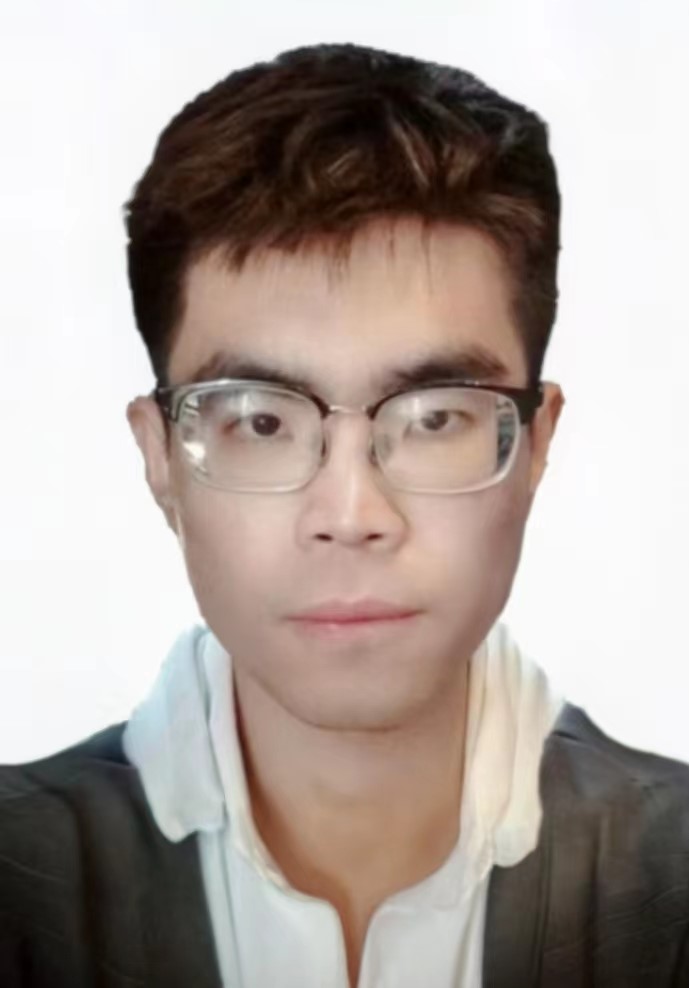}}]{Yinan Deng} received the B.S. degree from Beijing Institute of Technology, Beijing, China, in 2021. He is currently working toward the Ph.D. degree for the first year in intelligent navigation with School of Automation, Beijing Institute of Technology, Beijing, China. His research interests include 3D reconstruction and neural radiance fields.
 \end{IEEEbiography}

 \begin{IEEEbiography}[{\includegraphics[width=1in,height=1.25in,clip,keepaspectratio]{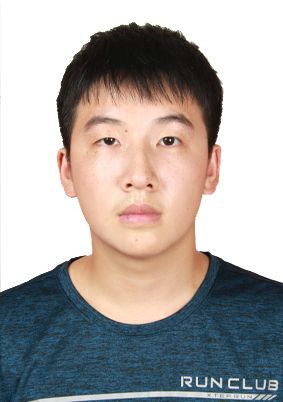}}]{Jiahui Wang} received the B.S. degree from Central South University, Changsha, China, in 2023. He is currently working toward the M.Sc. degree for the first year in intelligent navigation with School of Automation, Beijing Institute of Technology, Beijing, China. His research interests include neural implicit mapping and open-vocabulary mapping of robotic systems in practical environments.
 \end{IEEEbiography}

\begin{IEEEbiography}[{\includegraphics[width=1in,height=1.25in,clip,keepaspectratio]{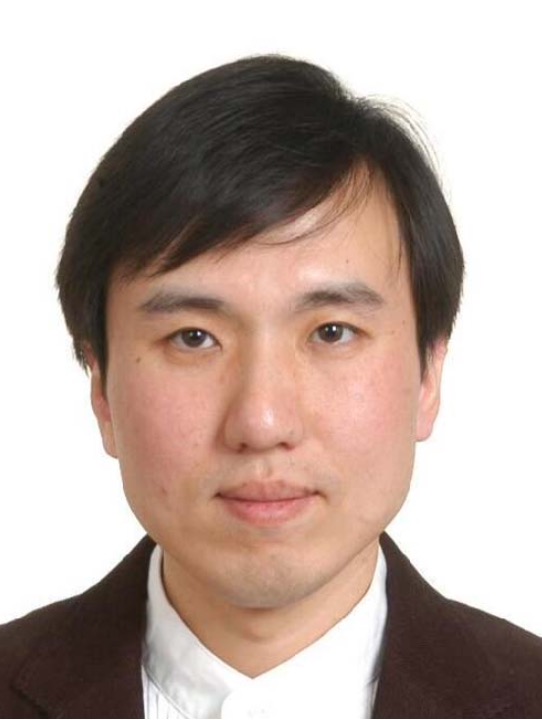}}]{Yi Yang}
received the Ph.D. degree in automation from the Beijing Institute of Technology, Beijing, China, in 2010. He is currently a Professor at the School of Automation, Beijing Institute of Technology. His research interests include autonomous vehicles, bioinspired robots, intelligent navigation, semantic mapping, scene understanding, motion planning and control. He is the author or co-author of more than 50 conference and journal papers in the area of unmanned ground vehicles.
\end{IEEEbiography}

\vfill

\end{document}